\pdfoutput=1

\documentclass[11pt]{article}

\usepackage[]{EMNLP2022}

\usepackage{times}
\usepackage{latexsym}
\usepackage{float}

\usepackage[T1]{fontenc}

\usepackage[utf8]{inputenc}

\usepackage{microtype}

\usepackage{amsfonts}
\usepackage{amsmath}
\usepackage{cleveref}
\usepackage{multirow}
\usepackage{adjustbox}
\usepackage{booktabs}
\usepackage{subcaption}     
\usepackage{enumitem}

\usepackage{amssymb}
\usepackage{pifont}

\newcommand\blfootnote[1]{%
  \begingroup
  \renewcommand\thefootnote{}\footnote{#1}%
  \addtocounter{footnote}{-1}%
  \endgroup
}

\title{Exploring Dual Encoder Architectures for Question Answering}

\author{
    Zhe Dong$^\dagger$
    \quad Jianmo Ni$^\dagger$
    \quad Daniel M. Bikel$^\ddagger$\thanks{\ \ Work done while at Google.}
    \quad Enrique Alfonseca$^\dagger$ \\
    \quad \textbf{Yuan Wang$^\dagger$}
    \quad \textbf{Chen Qu$^\dagger$}
    \quad \textbf{Imed Zitouni$^\dagger$}\\
    $^\dagger$Google Inc $^\ddagger$Meta AI\\
    \texttt{\small \{zhedong, jianmon, ealfonseca, yuawang, cqu, izitouni\}@google.com, dbikel@meta.com}
}

\begin{document}
\maketitle

\begin{abstract}
Dual encoders have been used for question-answering (QA) and information retrieval (IR) tasks with good results. Previous research focuses on two major types of dual encoders, Siamese Dual Encoder (SDE), with parameters shared across two encoders, and Asymmetric Dual Encoder (ADE), with two distinctly parameterized encoders. In this work, we explore different ways in which the dual encoder can be structured, and evaluate how these differences can affect their efficacy in terms of QA retrieval tasks. By evaluating on MS MARCO, open domain NQ and the MultiReQA benchmarks, we show that SDE performs significantly better than ADE. We further propose three different improved versions of ADEs by sharing or freezing parts of the architectures between two encoder towers. We find that sharing parameters in projection layers would enable ADEs to perform competitively with or outperform SDEs. We further explore and explain why parameter sharing in projection layer significantly improves the efficacy of the dual encoders, by directly probing the embedding spaces of the two encoder towers with t-SNE algorithm \citep{vandermaaten2008tsne}.
\blfootnote{Code and additional information can be found in: \url{https://sites.google.com/view/explore-dual-encoder-architect}.}
\end{abstract}

\section{Introduction}

\begin{figure*}[t]
  \centering
  \includegraphics[width=\linewidth]{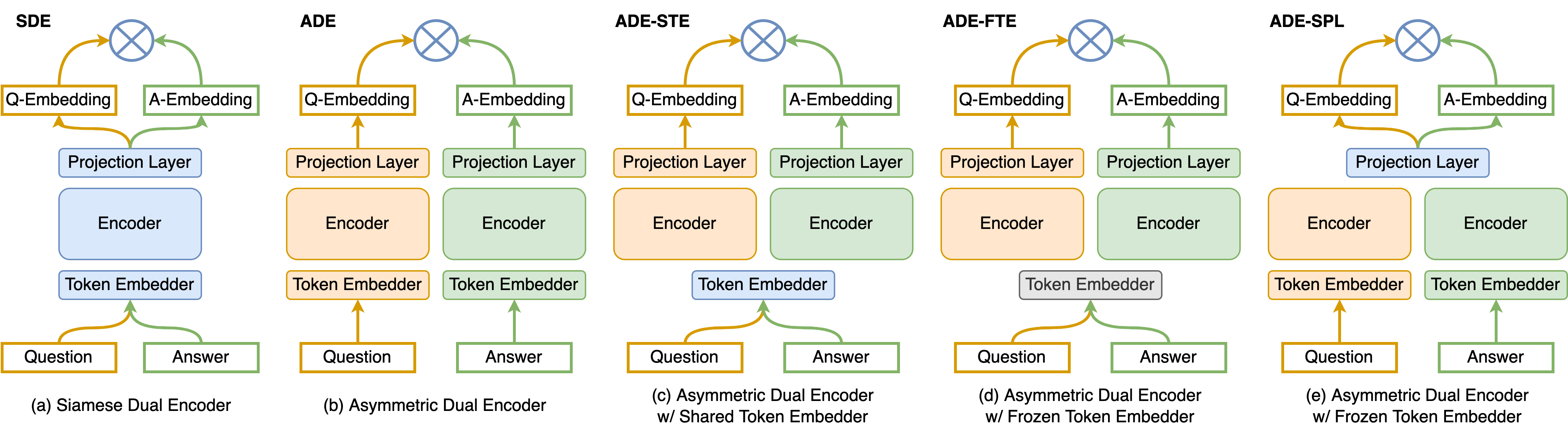}
  \caption{\footnotesize
  Architectures of dual encoders. We study whether parameter sharing in different dual encoder components (i.e.~token embedder, transformer encoder, and projection layer) can lead to better representation learning. {\color{orange}\textbf{Orange}} and {\color{teal}\textbf{green}} components are distinctly parameterized for question and answer encoder towers, respectively. {\color{blue}\textbf{Blue}} components are shared between two encoding paths. {\color{gray}\textbf{Grey}} components are frozen during the fine-tuning.} 
  \label{fig:de}
  \vspace{-1em}
\end{figure*}

A dual encoder is an architecture consisting of two encoders, each of which encodes an input (such as a piece of text) into an embedding, and where the model is optimized based on similarity metrics in the embedding space. It has shown an excellent performance in a wide range of information retrieval and question answering tasks \citep{Gillick2018EndtoEndRI,Karpukhin2020DensePR}. This approach is also easy to productionize because the embedding index of dual encoders can grow dynamically for newly discovered or updated documents and passages without retraining the encoders \citep{Gillick2018EndtoEndRI}. In contrast, generative neural networks used for question answering need to be retrained with new data. This advantage makes dual encoders more robust to freshness. 

\begin{table}[t]
\small
    \centering
    \begin{tabular}{l|r}
    \toprule
    Model & Architecture \\
    \midrule
    DPR \citep{Karpukhin2020DensePR} & Asymmetric  \\
    DensePhrases \citep{Lee2021LearningDR} & Asymmetric  \\
    SBERT \citep{Reimers2019SentenceBERTSE} & Siamese  \\
    ST5 \citep{jianmo2021sentencet5}  & Siamese  \\ 
    \bottomrule
    \end{tabular}
    \caption{Existing off-the-shelf dual encoders.}
    \label{tab:off-the-shelf}
    \vspace{-1em}
\end{table}

There are different valid designs for dual encoders. As shown in \Cref{tab:off-the-shelf}, previous research focuses on the two major types: Siamese Dual Encoder (SDE) and Asymmetric Dual Encoder (ADE). In a SDE, the parameters are shared between the two encoders. In an ADE, only some or no parameters are shared \citep{Gillick2018EndtoEndRI}. In practice, we often require certain asymmetry in the dual encoders, especially in the case where the inputs of the two towers are of different types. Although all of these models have achieved excellent results in different NLP applications, how these parameter-sharing design choices affect the model performance is largely unexplored.

This paper explores the impact of parameter sharing in \textit{different components} of dual encoders on question answering tasks, and assess whether the impact holds for dual encoders with different model capacity. In particular, we compare five different variants of dual encoders as shown in \Cref{fig:de}:
\begin{itemize}[noitemsep,topsep=2pt,parsep=2pt,partopsep=2pt,leftmargin=6mm]
    \item Siamese Dual-Encoder (SDE),
    \item Asymmetric Dual-Encoder (ADE),
    \item ADE with shared token embedder (ADE-STE),
    \item ADE with frozen token embedder (ADE-FTE),
    \item ADE with shared projection layer (ADE-SPL),
\end{itemize}
where the two extreme cases are SDE and ADE, with the parameters of two towers completely shared or distinct.

We conduct experiments across $7$ well-established datasets. We find that SDEs consistently outperforms ADEs on question answering retrieval tasks, and sharing parameters in token embedders and projection layers between the two encoders improves the efficacy of ADEs.  In particular, sharing projection layer (ADE-SPL) enables ADEs to achieve comparable or even better performance than SDEs.

To better understand why parameter sharing improves the efficacy of the asymmetric dual encoders, we directly analyze the embeddings from the two encoder tower, by projecting and cluster them into $2$-dimensional space using t-SNE \citep{vandermaaten2008tsne}. The analysis shows that without sharing projection layer, ADEs tend to embed the inputs of the two encoder towers into disjoint embedding spaces, which hinders the quality of retrieval. Based on the findings, we recommend to share the projection layers between two encoder towers in practice, if using asymmetric dual encoder is necessary.

\section{Related work}

Dual encoders have been widely studied in entity linking \citep{Gillick2018EndtoEndRI}, open-domain question answering \citep{Karpukhin2020DensePR}, and dense retrieval \citep{Ni2021LargeDE}, etc. This architecture consists of two encoders, where each encoder encodes arbitrary inputs that may differ in type or granularity, such as queries, images, answers, passages, or documents.

Open-domain question answering (ODQA) is a challenging task that searches for evidence across large-scale corpora and provides answers to user queries \citep{Voorhees99thetrec-8,Chen2017ReadingWT}. One of the prevalent paradigms for ODQA is a two-step approach, consisting of a retriever to find relevant evidence and a reader to synthesize answers. Alternative approaches are directly retrieving from large candidate corpus to provide sentence-level \citep{guo-etal-2021-multireqa} or phrase-level \citep{lee2021learning} answers; or directly generating answers or passage indices using an end-to-end generation approach \citep{Tay2022TransformerMA}. \citet{Lee2021LearningDR} compared the performance of SDEs and ADEs for phrase-level QA retrieval tasks. However, they only considered the two extreme cases, where two towers have the parameters completely shared or distinct. In this work, we address the missing piece of previous work by exploring parameter sharing in different parts of the model.

\begin{table*}[!t]
\centering
\adjustbox{max width=\textwidth}{
\small
\begin{tabular}{cccccccc}
\toprule
\textbf{Metric} & \textbf{Model}
    & \textbf{MSMARCO}  & \textbf{NQ}       & \textbf{SQuAD}
    & \textbf{TriviaQA} & \textbf{HotpotQA} & \textbf{SearchQA} \\
\midrule
\multirow{7}{*}{P@1} & SDE 
    & $\mathbf{15.92}$    & $48.87$    & $\mathbf{70.13}$
    & $36.55$    & $\mathbf{34.36}$    & $36.40$ \\
        & ADE
    & $14.20$    & $47.83$    & $60.39$
    & $31.30$    & $26.71$    & $39.48$ \\
        & ADE-STE
    & $14.71$    & $48.29$    & $61.05$
    & $33.59$    & $28.71$    & $40.43$ \\
        & ADE-FTE
    & $14.23$    & $49.38$    & $62.86$
    & $35.11$    & $29.07$    & $\mathbf{42.06}$ \\
        & ADE-SPL
    & $15.46$    & $\mathbf{50.06}$    & $69.39$
    & $\mathbf{38.17}$    & $33.66$    & $41.13$ \\
        & BERT-DE
    & -    & $36.22$    & $55.13$
    & $29.11$    & $32.05$    & $30.2$ \\
        & USE-QA
    & -    & $38$    & $66.83$
    & $32.58$    & $31.71$    & $31.45$ \\
\midrule
\multirow{7}{*}{MRR} & SDE
    & $\mathbf{28.49}$    & $61.15$    & $\mathbf{78.44}$
    & $49.29$    & $45.58$    & $54.26$ \\
        & ADE
    & $26.31$    & $59.38$    & $70.33$
    & $43.42$    & $37.27$    & $55.02$ \\
        & ADE-STE
    & $26.78$    & $59.81$    & $70.85$
    & $45.79$    & $39.14$    & $56.08$ \\
        & ADE-FTE
    & $26.64$    & $61.23$    & $72.18$
    & $46.95$    & $39.72$    & $57.44$ \\
        & ADE-SPL
    & $28.20$    & $\mathbf{61.92}$    & $77.65$
    & $\mathbf{50.3}$    & $44.19$    & $\mathbf{57.48}$ \\
        & BERT-DE
    & -    & $52.02$    & $64.74$
    & $41.34$    & $\mathbf{46.21}$    & $47.08$ \\
        & USE-QA
    & -    & $52.27$    & $75.86$
    & $42.39$    & $43.77$    & $50.7$ \\
\bottomrule
\end{tabular}
}
\caption{\label{tab:experiment-main} 
\footnotesize
Precision at $1$(P@1)($\%$) and Mean Reciprocal Rank (MRR)($\%$) on QA retrieval tasks. \texttt{SDE} and \texttt{ADE} stand for Siamese Dual-Encoder and Asymmetric Dual-Encoder, respectively. \texttt{ADE-STE}, \texttt{-FTE} and \texttt{-SPL} are the ADEs with shared token-embedders, frozen token-embedders, and shared projection-layers, respectively. \texttt{BERT-DE}, which stands BERT \citep{devlin-etal-2019-bert} Dual-Encoder, and \texttt{USE-QA} \citep{yang-etal-2020-multilingual} are the baselines reported in MultiReQA \citep{guo-etal-2021-multireqa}. The most performant models are marked in bold.  
}
\vspace{-1em}
\end{table*}

\section{Method}
\label{sec:method}

We follow a standard setup of QA retrieval: given a question $q$ and a corpus of answer candidates $\mathcal{A}$, the goal is to retrieve $k$ relevant answers $\mathcal{A}_k \in \mathcal{A}$ for $q$. The answer can be either a passage, a sentence, or a phrase. 

We adopt a \textit{dual encoder} architecture \citep{Gillick2018EndtoEndRI,Reimers2019SentenceBERTSE} as the model to match query and answers for retrieval. The model has two encoders, where each is a transformer that encodes a question or an answer. Each encoder first produces a fixed-length representation for its input and then applies a projection layer to generate the final embedding. 

We train the dual encoder model by optimizing the contrastive loss with an in-batch sampled softmax~\citep{Henderson2017EfficientNL}:
\begin{equation}
    \mathcal{L} = \frac{e^{\text{sim}(q_i, a_i)/ \tau}}{\sum_{j \in \mathcal{B}} { e^{\text{sim}(q_i, a_j) / \tau} } },
    \label{eq::loss}
\end{equation}
where $q_i$ is a question and $a_*$ is a candidate answer. $a_i$ is ground-truth answer, or a positive sample, for $q_i$. All other answers $a_j$ in the same batch $\mathcal{B}$ are considered as negative samples during training. $\tau$ is the softmax temperature and $\mathbf{sim}$ is a similarity function to measure the relevance between the question and the answer. In this work, we use cosine distance as the similarity function:
\begin{equation}
    \text{sim}(q_i, a_j) = 
    \frac{\vec{q_i}\cdot \vec{a_j}}
    {\lVert \vec{q_i} \rVert \lVert \vec{a_j} \rVert}.
\end{equation}

\section{Experiments and Analysis}
We evaluate the proposed dual encoder architectures on six question-answering retrieval tasks from MS MARCO \citep{bajaj2016msmarco} and MultiReQA \citep{guo-etal-2021-multireqa}. In MS MARCO, we consider the relevant passages as answer candidates, while for the five QA datasets in MultiReQA the answer candidates are individual sentences. We further validate the conclusion on an open domain question-answering task, Open Domain NaturalQuestions, where the retrieval candidates are context passages.

To initialize the parameters of dual encoders, we use pre-trained \texttt{t5.1.1} encoders \citep{colin2020t5}. Following \citet{jianmo2021sentencet5}, we take the average embeddings of the T5 encoder's outputs and send to a projection layer to get the final embeddings. The projection layers are randomly initialized, with variance scaling initialization with scale $1.0$. For the retrieval, we use mean embeddings from the encoder towers. To make a fair comparison, the same hyper-parameters are applied across all the models for the fine-tuning with Adafactor optimizer \citep{shazeer2018adafactor}, using learning rate $10^{-3}$ and batch size $512$. The models are fine-tuned for $20,000$ steps, with linear decay of learning rate from $10^{-3}$ to $0$ at the final steps. The fine-tuned models are benchmarked with precision at $1$ (P@1) and mean reciprocal rank (MRR) on the QA retrieval tasks, in \Cref{tab:experiment-main}.

\begin{table}
\centering
\adjustbox{max width=\linewidth}{
\small
\begin{tabular}{c|ccccc|cc}
\toprule
\textbf{Model}  & SDE & ADE & A-STE & A-FTE & A-SPL & D-G7 & D-G127 \\
\midrule
Top-5   & $62.2$ & $57.6$ & $58.0$ & $57.4$ & $\mathbf{62.7}$ & $51.1$ & $55.8$ \\
Top-20  & $\mathbf{77.0}$ & $73.2$ & $73.1$ & $73.8$ & $76.4$ & $69.1$ & $73.0$ \\
Top-100 & $\mathbf{84.6}$ & $82.7$ & $82.5$ & $83.2$ & $84.4$ & $80.8$ & $83.1$ \\
\bottomrule
\end{tabular}
}
\caption{\label{tab:experiment-opennq} 
\footnotesize
Evaluation of different dual encoders,
measured as top-k retrieval accuracy on Open Domain Natural Questions (development set). The baselines are quoted from DRP \citep{Karpukhin2020DensePR} with golden labels and $7$ (D-G7) or $127$ (D-G127) negative examples.
}
\vspace{-1em}
\end{table}

\begin{figure*}
    \centering
    \begin{subfigure}{0.19\textwidth}
        \centering
        \includegraphics[width=\linewidth]{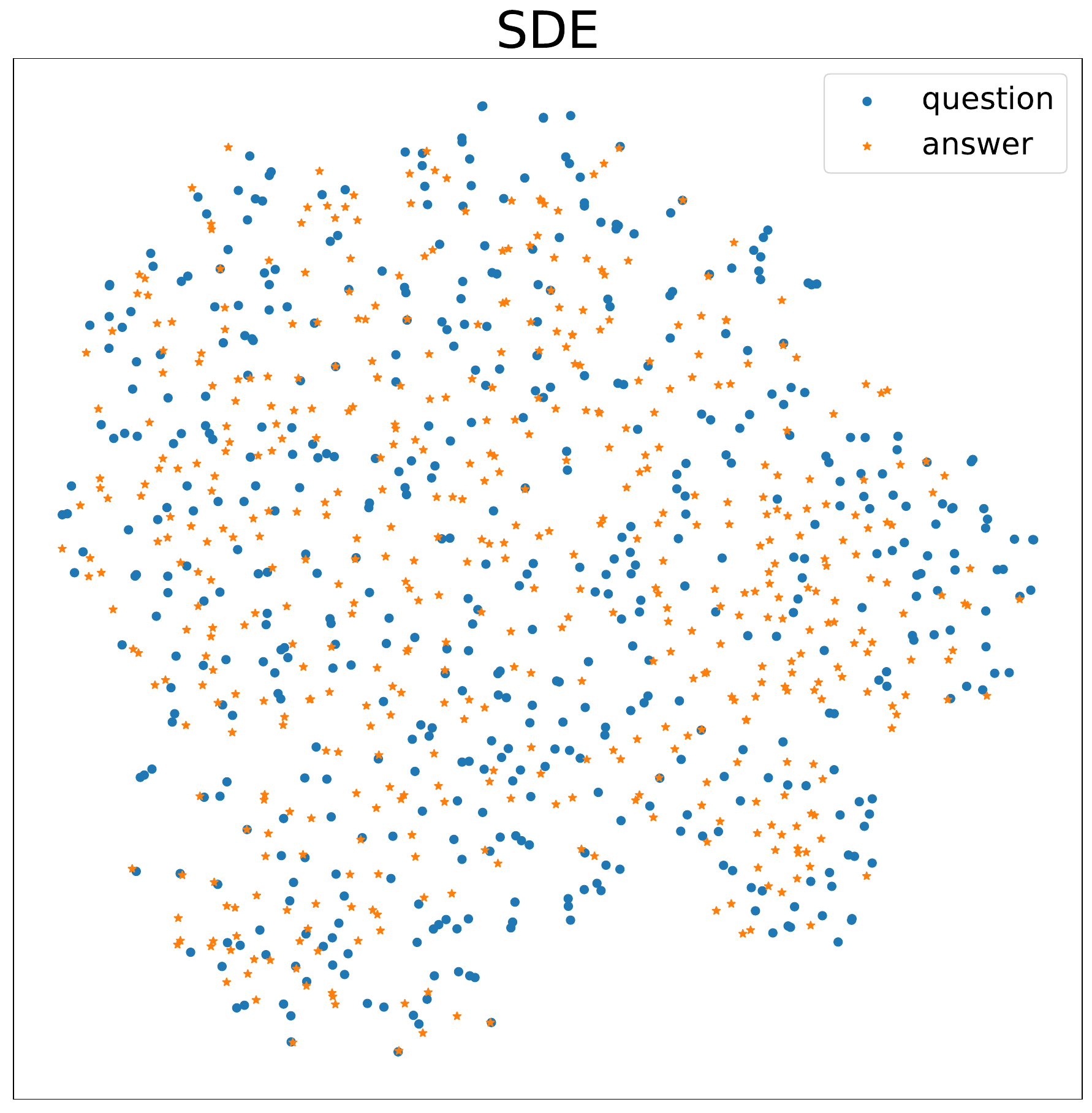}
    \end{subfigure}
    \begin{subfigure}{0.19\textwidth}
        \centering
        \includegraphics[width=\linewidth]{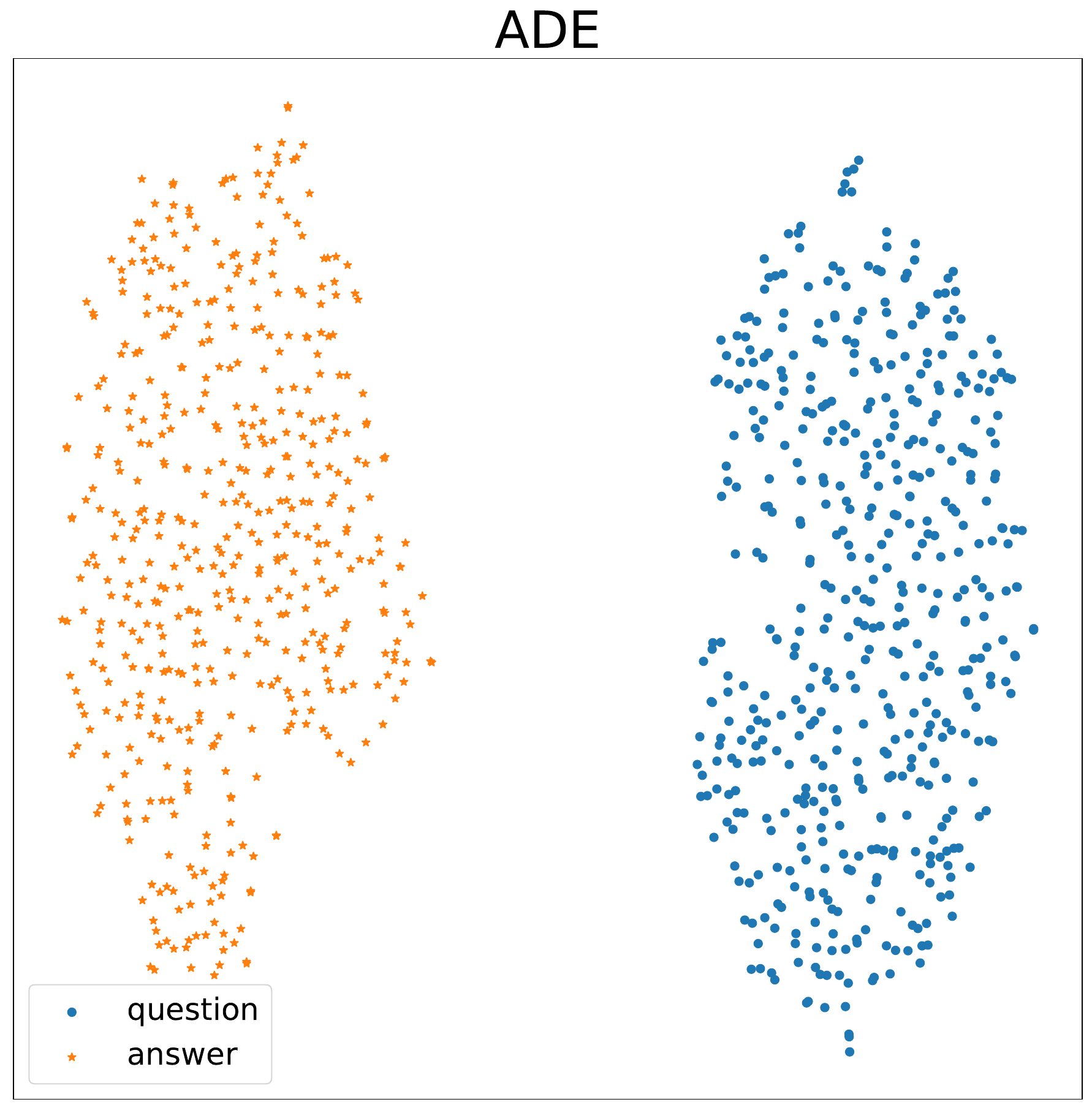}
    \end{subfigure}
    \begin{subfigure}{0.19\textwidth}
        \centering
        \includegraphics[width=\linewidth]{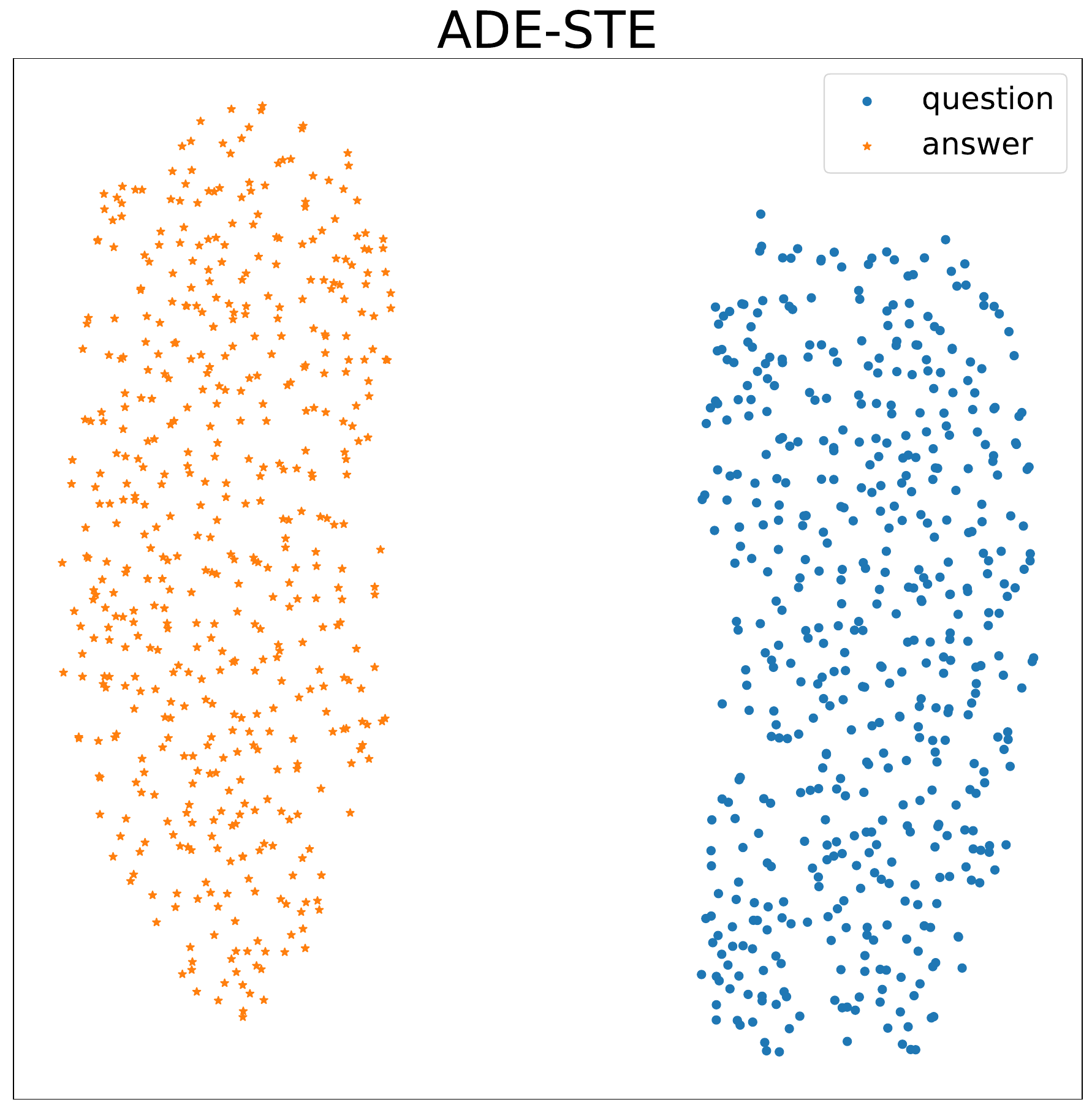}
    \end{subfigure}
    \begin{subfigure}{0.19\textwidth}
        \centering
        \includegraphics[width=\linewidth]{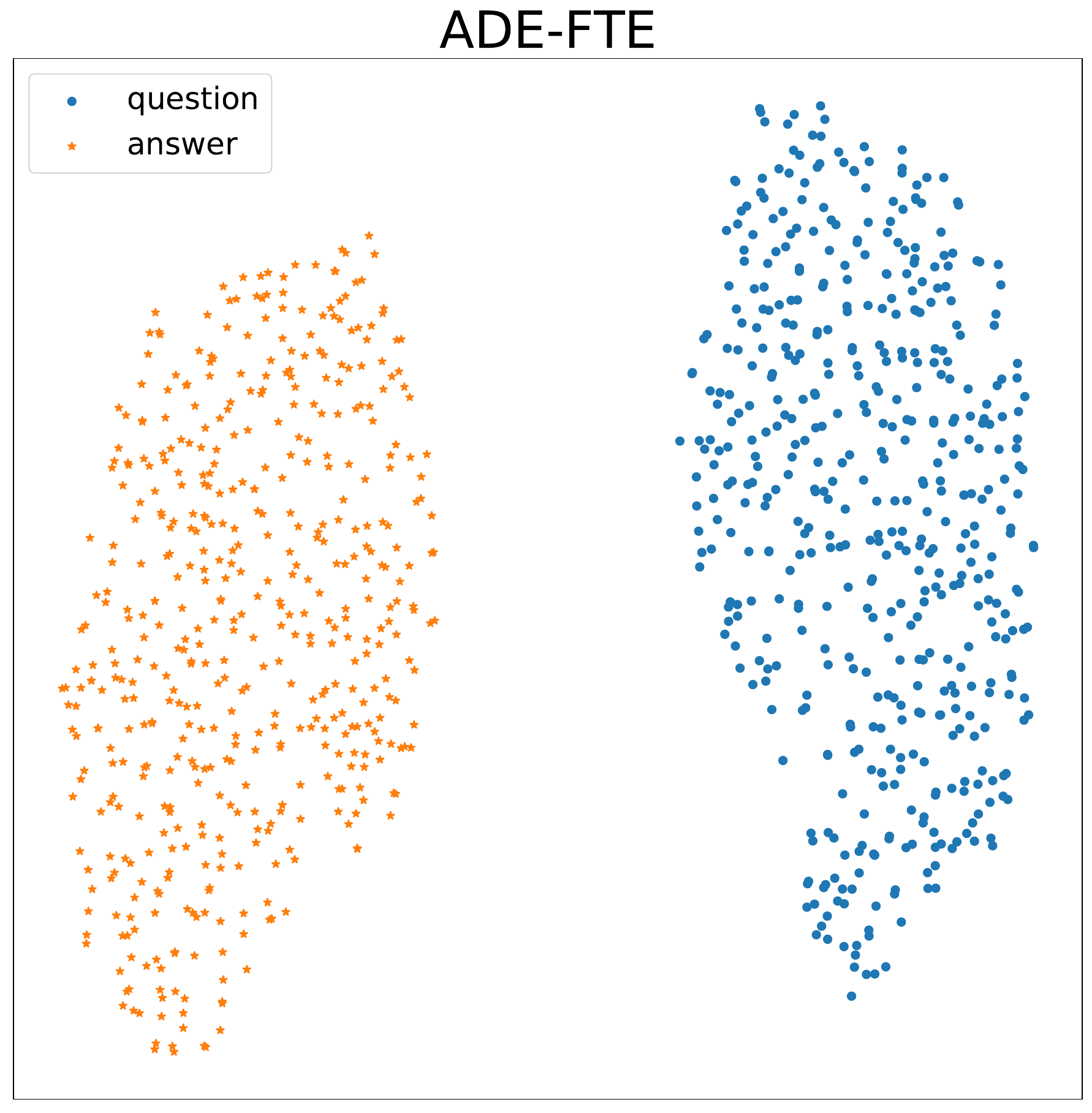}
    \end{subfigure}
    \begin{subfigure}{0.19\textwidth}
        \centering
        \includegraphics[width=\linewidth]{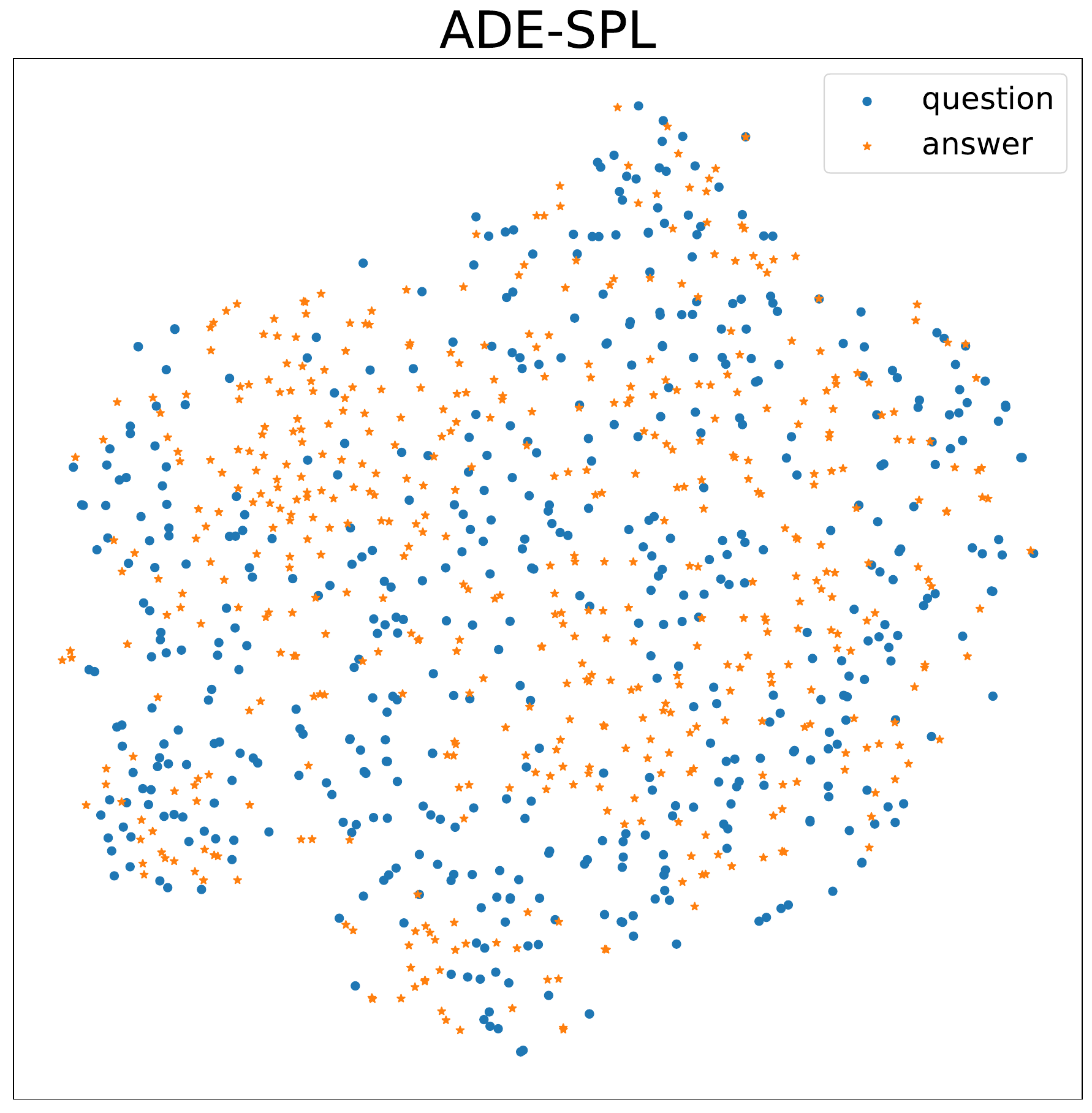}
    \end{subfigure}
    \caption{
    \footnotesize
    t-SNE clustering of the embeddings of the NaturalQuestions eval set generated by five dual encoders.
    The blue and orange points represent the embeddings of the questions and answers, respectively.
    \label{figure:embedding-tsne}}
\end{figure*}

\subsection{Comparing SDE and ADE}
SDE and ADE in \Cref{fig:de} (a) and (b) are the two most distinct dual-encoders in terms of parameter sharing. 
Experiment results show that, on QA retrieval tasks, ADE performs consistently worse than SDE. To explain that, our \textbf{assumption} is that, at inference time, the two distinct encoders in ADE that do not share any parameters, map the questions and the answers into two parameter spaces that are not perfectly aligned. However, for SDE, parameter sharing enforces the embeddings from the two encoders to be in the same space. We verify this assumption in \Cref{sec:analysis-on-embeddings}.

\subsection{Improving the Asymmetric Dual Encoder}
Although the dual encoders with maximal parameter sharing (SDEs) performs significantly better than the ones without parameter sharing (ADEs), we often require certain asymmetry in the dual encoders in practice. 
Therefore, trying to improve the performance of ADEs, we construct dual-encoders with parameters shared at different levels between the two encoders. 

\paragraph{Shared and Frozen Token Embedders.} 
Token embedders are the lowest layers close to the input text. In ADEs, token embedders are initialized from the same set of pre-trained parameters, but fine-tuned separately. A straightforward way to bring ADEs closer to SDEs is to share the token embedders between the two towers, or to an extreme, to simply freeze the token embedders during training.

Evaluated on MS MARCO and MultiReQA, the results in \Cref{tab:experiment-main} show that both freezing (ADE-FTE) and sharing (ADE-STE) token embedders bring consistent, albeit marginal, improvements for ADEs. However, ADEs with common token embedders still leave a significant gap compared to SDEs on most tasks. These results suggest token embedders might not be the key to close this gap.

\begin{figure}[t]
\centering
\includegraphics[width=0.8\columnwidth]{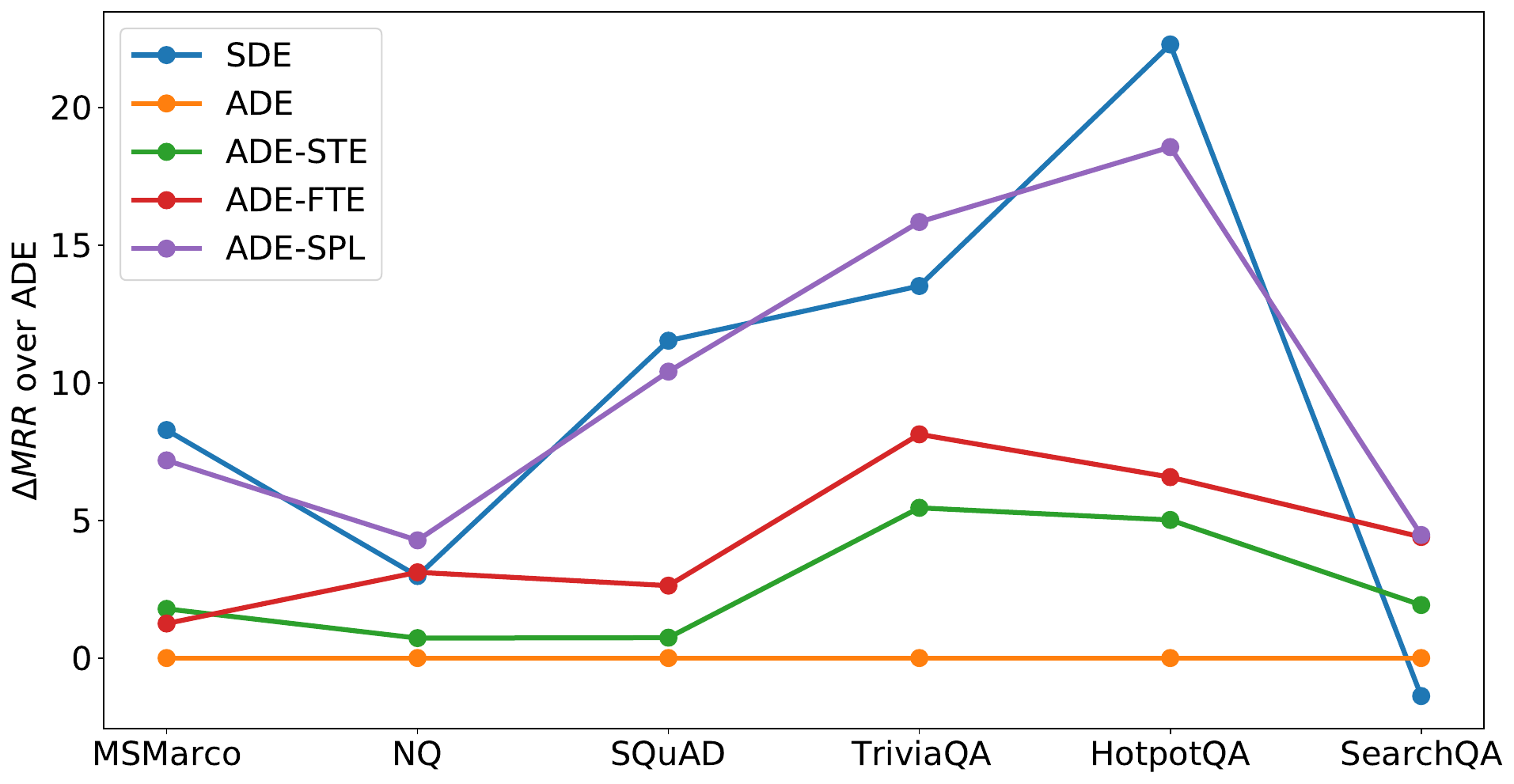}
\caption{\label{figure:delta-mrr-main}
\footnotesize
Relative performance improvements of different models relative to ADE on QA retrieval tasks. $\Delta \mathrm{MRR} = (\mathrm{MRR} - \mathrm{MRR}_{ADE})/\mathrm{MRR}_{ADE}) \times 100$.
}
\end{figure}

\paragraph{Shared Projection Layers.}
Another way of improving retrieval quality of ADEs is to share the projection layers between the two encoders. \Cref{tab:experiment-main} shows that sharing projection layers drastically improves the quality of ADEs. As in \Cref{figure:delta-mrr-main}, ADE-SPL (purple curve) performs on-par and, sometimes, even better than SDE (blue curve). This observation reveals that sharing projection layers is a valid approach to enhance the performance of ADEs. This technique would be vital if asymmetry is required by a modeling task.

\subsection{Analysis on the Embeddings}
\label{sec:analysis-on-embeddings}
The experiments corroborate our assumption that sharing the projection layer enforces the two encoders to produce embeddings in the same parameter space, which improves the retrieval quality.

To further substantiate our assumption, we first generate the question and answer embeddings from the NaturalQuestions eval set, and then use t-SNE \citep{vandermaaten2008tsne} to project and cluster the embeddings into $2$-dimensional space.\footnote{For efficiently clustering with t-SNE, we randomly sampled questions and answers, $400$ each, from the NQ eval set.} \Cref{figure:embedding-tsne} shows that, for ADE, ADE-STE and ADE-FTE that have separate projection layers, the question and answer embeddings are projected and clustered into two disjoint groups. In comparison, ADE-SPL that shares the projection layers, the embeddings of questions and answers are not separable by t-SNE, which is similar to the behavior of SDE.
This verifies our assumption that the projection layer plays an important role in bringing together the representations of questions and answers, and is the key for retrieval performance. 

\subsection{Experiment on Open Domain NQ.} 
To further validate our assumption on ODQA, \Cref{tab:experiment-opennq} shows the comparison for different dual encoder architectures and the baselines from DRP \citep{Karpukhin2020DensePR} on Open Domain Natural Questions (OpenNQ) dataset, using the top-$k$ accuracy ($k \in \{5, 20, 100\}$). SDE and ADE-SPL perform competitively on the OpenQA passage retrieval task.  

\begin{figure}[t]
\centering
    \includegraphics[width=0.47\linewidth]{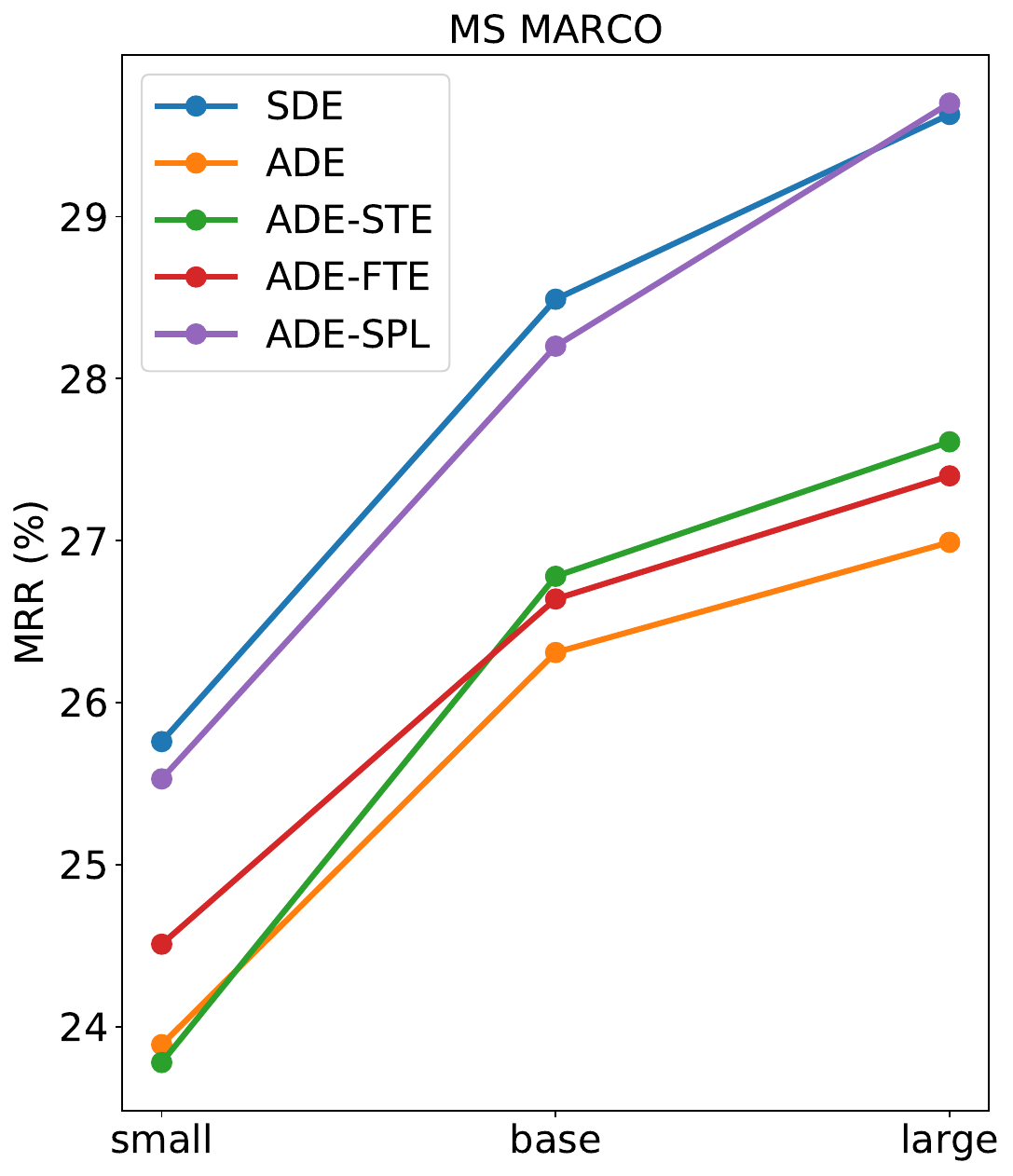}
    \includegraphics[width=0.49\linewidth]{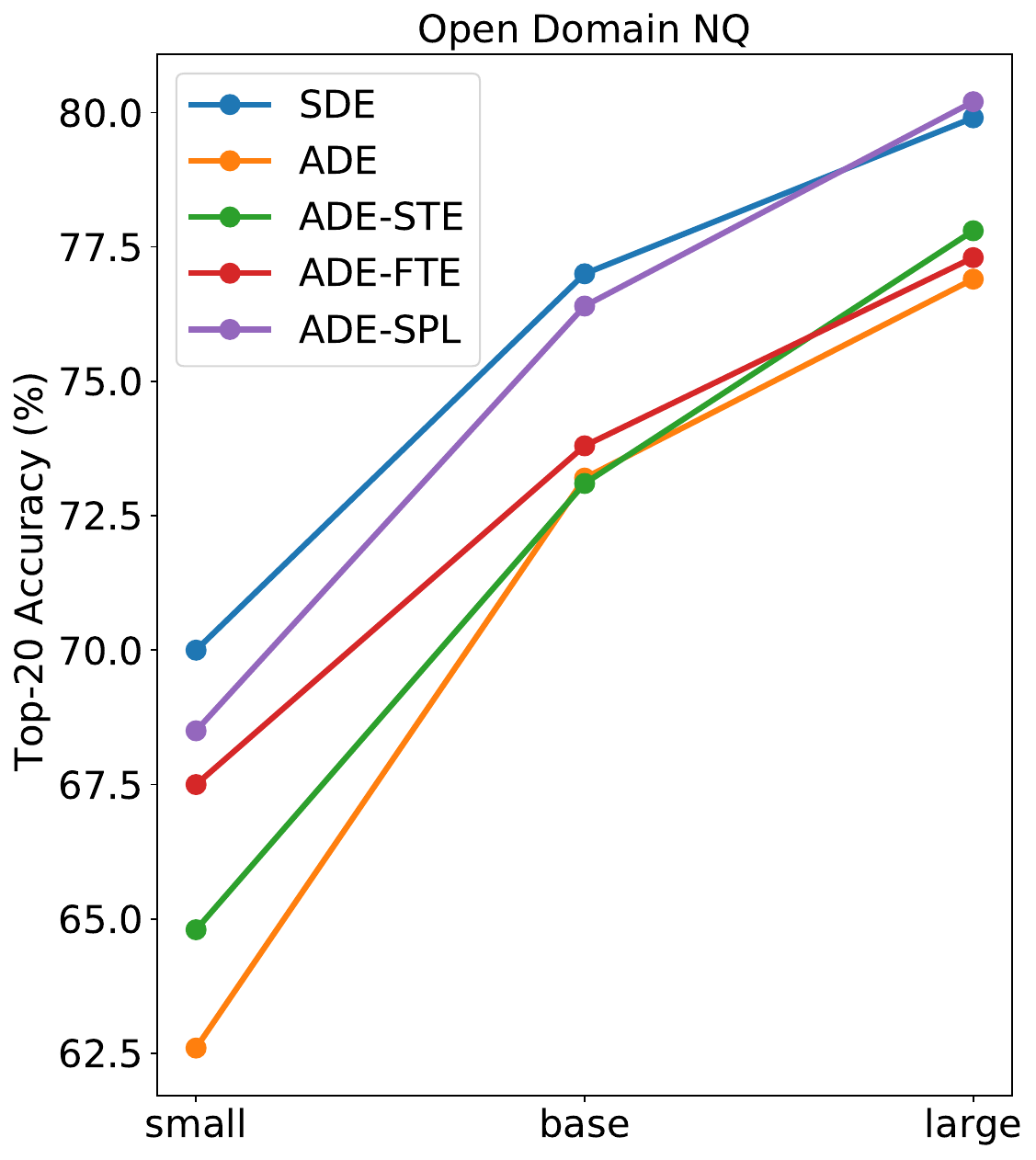}
\caption{\label{figure:mrr-scaling}
\footnotesize
The impact of model size on the performance of different dual encoder architectures, measured by MRR on the eval set of MS MARCO (left), and Top-20 Accuracy on development set of Open Domain NQ (right). 
}
\end{figure}

\begin{table*}[ht]
\centering
\adjustbox{max width=\textwidth}{
\small
\begin{tabular}{c|ccc|ccc|ccc}
\toprule
\textbf{Metric} & \multicolumn{3}{c}{Top-5} & \multicolumn{3}{c}{Top-20} & \multicolumn{3}{c}{Top-100} \\
\midrule
\textbf{Model Size}  & \texttt{small} & \texttt{base} & \texttt{large} & \texttt{small} & \texttt{base} & \texttt{large} & \texttt{small} & \texttt{base} & \texttt{large} \\
\midrule
SDE	& $\mathbf{54.63}$	& $62.24$	& $67.51$	& $\mathbf{70.00}$	& $\mathbf{76.98}$	& $79.92$	& $\mathbf{80.28}$	& $\mathbf{84.57}$	& $\mathbf{86.93}$ \\
ADE	& $46.32$	& $57.65$	& $62.88$	& $62.58$	& $73.21$	& $76.93$	& $74.99$	& $82.69$	& $85.35$ \\
ADE-STE	& $48.03$	& $58.03$	& $64.04$	& $64.85$	& $73.07$	& $77.84$	& $76.57$	& $82.55$	& $85.21$ \\	
ADE-FTE	& $50.50$	& $57.40$	& $63.74$	& $67.51$	& $73.80$	& $77.26$	& $78.45$	& $83.19$	& $85.32$ \\
ADE-SPL	& $53.52$	& $\mathbf{62.66}$	& $\mathbf{68.06}$	& $68.48$	& $76.37$	& $\mathbf{80.25}$	& $79.11$	& $84.40$	& $86.23$ \\
\bottomrule
\end{tabular}
}
\caption{\label{tab:experiment-scaling-opennq} 
Evaluation of the scaling effect on Open Domain Natural Questions, using top-k retrieval accuracy, with dual encoders initialized from \texttt{t5.1.1-small}, \texttt{-base}, and \texttt{-large} checkpoints. The most performant models are marked in bold.
}
\end{table*}

\begin{table}[!h]
\centering
\adjustbox{max width=\linewidth}{
\begin{tabular}{c|ccc|ccc}
\toprule
\textbf{Metric} & \multicolumn{3}{c}{P@1} & \multicolumn{3}{c}{MRR} \\
\midrule
\textbf{Model Size}  & \texttt{small} & \texttt{base} & \texttt{large} & \texttt{small} & \texttt{base} & \texttt{large} \\
\midrule
SDE     & $\mathbf{14.50}$ & $\mathbf{15.92}$ & $\mathbf{16.53}$ & $\mathbf{25.76}$ & $\mathbf{28.49}$ & $29.63$  \\
ADE     & $13.31$ & $14.20$ & $14.17$ & $23.89$ & $26.31$ & $26.99$  \\
ADE-STE & $12.99$ & $14.71$ & $15.14$ & $23.78$ & $26.78$ & $27.61$  \\
ADE-FTE & $13.67$ & $14.23$ & $14.73$ & $24.51$ & $26.64$ & $27.40$  \\
ADE-SPL & $14.31$ & $15.46$ & $16.42$ & $25.53$ & $28.20$ & $\mathbf{29.70}$  \\
\bottomrule
\end{tabular}
}
\caption{\label{tab:experiment-scaling-msmarco} 
Evaluation of the scaling effect on MS MARCO \citep{bajaj2016msmarco} QA retrieval tasks, using Precision at $1$ (P@1)($\%$) and Mean Reciprocal Rank (MRR)($\%$), with dual encoders initialized from \texttt{t5.1.1-small}, \texttt{-base}, and \texttt{-large} checkpoints. The most performant models are marked in bold.
}
\end{table}

\subsection{Impact of Model Size.} 

To assess the impact of model size, we fine-tune and evaluate the dual-encoders initialized from \texttt{t5.1.1-small} ($\sim77$M parameters), \texttt{-base} ($\sim250$M), and \texttt{-large} ($\sim800$M) on the MS MARCO and OpenNQ. \Cref{figure:mrr-scaling}, and \Cref{tab:experiment-scaling-opennq} and \ref{tab:experiment-scaling-msmarco}, show that, across different model sizes, sharing projection layers consistently improves the retrieval performance of ADE, and ADE-SPL performs competitively with SDE. 
This observation further validates our recommendation to share the projection layer in ADEs.

\section{Conclusion and Future Work}
Based on the experiments on $7$ QA retrieval tasks with three different model sizes, we conclude that, although SDEs outperforms ADEs, sharing the projection layer between the two encoders enables ADEs to perform competitively with SDEs. By directly probing the embedding space, we demonstrate that the shared projection layers in SDE and ADE-SPL maps the embeddings of the two encoder towers into coinciding parameter spaces, which is crucial for improving the retrieval quality. Therefore, we \textbf{recommend} to share the projection layers between two encoders of ADEs in practice.

\section{Limitations}
In our work, we focus on dual encoder architectures and their impacts on the QA retrieval quality. Other dense retrieval models beyond dual encoders, e.g. ColBERT \citep{Khattab2020ColBERTEA} and DensePhrases \citep{Lee2021LearningDR}, are beyond the scope of this work. In addition, we demonstrate that sharing projection layer is a simple yet effective technique to improve the quality of asymmetric dual encoder for QA retrieval tasks. However, other more complicated architectural improvements of ADE, e.g. introducing more complexity in the projection layers or interactions between two encoder towers, are beyond the scope of this work. 
\section*{Acknowledgements}
We could like to thank the anonymous reviewers of EMNLP 2022, who provided insightful feedback and constructive suggestions that significantly helped us on improving the writing of the paper. 

\bibliography{anthology,custom}
\bibliographystyle{acl_natbib}

\end{document}